# Optimal initialization of K-means using Particle Swarm Optimization

Ashutosh Mahesh Pednekar
*School of Computer Science & Engineering*
VIT University
Vellore, India
ashutoshpednekar15@gmail.com

*Abstract*— **This paper proposes the use of an optimization algorithm, namely PSO to decide the initial centroids in K-means, to eventually get better accuracy. The vecorized notation of the optimal centroids can be thought of as entities in an optimization space, where the accuracy of K-means over a random subset of the data could act as a fitness measure. The resultant optimal vector can be used as the initial centroids for K-means.**

*Keywords—centroid, optimization, vectorization, Swarm, sampling*

## I. INTRODUCTION

Unsupervised learning has always been a promising field of study, mainly because of the versatility of its applications. Cases where the class label is not available are often encountered in day to day life. Over the years, unsupervised learning has come to what we know today. K-means is one of the simplest yet very powerful algorithm in all of machine learning. From it's humble origins to the more advanced versions like K-means++, it has always had a special place in data science. Almost all these advancements were intended to better the performance of the algorithm by making intelligent choices while initializing the centroids.

An optimal initialization can almost all the time guarantee faster convergence and better accuracy over most datasets with localized data points. This paper puts forth the use of PSO optimization algorithm as an approach to choose the ideal centroids.

### A. K-means Clustering : Overview

Being the simplest, yet powerful clustering algorithm, Kmeans still holds its crown as the go to choice for most clustering problems. The underlying principle of K-means is to iteratively spawn clusters of data points according to their Euclidean distance from the current centroids. Once clusters are formed, new centroids are chosen by taking the average of all the points in a cluster.

This is repeated until the centroids converge to an optimal location.This can also be thought of as optimizing an 'Objective function' that best clusters the points in the data space.

### B. Particle Swarm Optimization: Overview

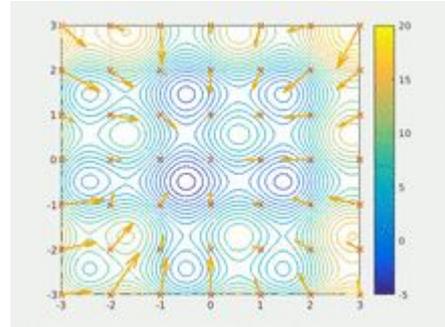

particle swarm optimization (PSO) is an optimization algorithm that computationally optimizes a given Loss surface by iteratively trying to improve a candidate solution with regard to a given measure of quality. It solves a problem by having a population of candidate solutions, here dubbed particles, and moving these particles around in the search-space according to simple mathematical formulae over the particle's position and velocity.

How each particle moves is influenced by its local best known position, but is also guided toward the best known positions in the search-space, which are updated as better positions are found by other particles. This is expected to move the swarm toward the best solutions. This approach of not only rushing towards the supposed Global minima, but also giving weightage to the local neighbourhood of each and every particle, greatly improves the Exploration without compromising on Exploitation.

## II. METHODOLOGY

Consider this, what if you choose the centroids that are pretty close to where the actual cluster centroids should be? Wouldn't it be great if we could achieve this ? This is exactly what this paper tries to achieve. There have been several successful attempts to try to achieve this, like K-means++, etc. Having seen the advantages of the PSO optimization algorithm, it is well worth it to give it a try. Swarm optimization techniques require particles scattered across a loss surface. These particles are in not the data points involved, something entirely different.

K-means clustering inherently means that we intend to create k clusters, and hence k centroids need to be initialized. Considering our data to be N dimensional, we could stack these centroids horizontally as follows: -

<x1,y1,z1,...,x2,y2,z2,..................xn,yn,zn>

This vectorized notation can further be used to create particles for PSO. Here, the x1,x2,....xn represent the X coordinates of the centroids, y1,y2,...yn denote the respective Y coordinates, and so on…As for the objective function, we can simply use the root of the average distances of all the points from these centroids. For faster computation, we can also do random sampling on the data to get a smaller subset to calculate this metric. If you have the resources, you could also use multiprocessing or threading techniques to scale up over clusters or powerful data clusters. For the following illustrations, I've used the Google Colab platform and it's free GPU acceleration for my computational needs.

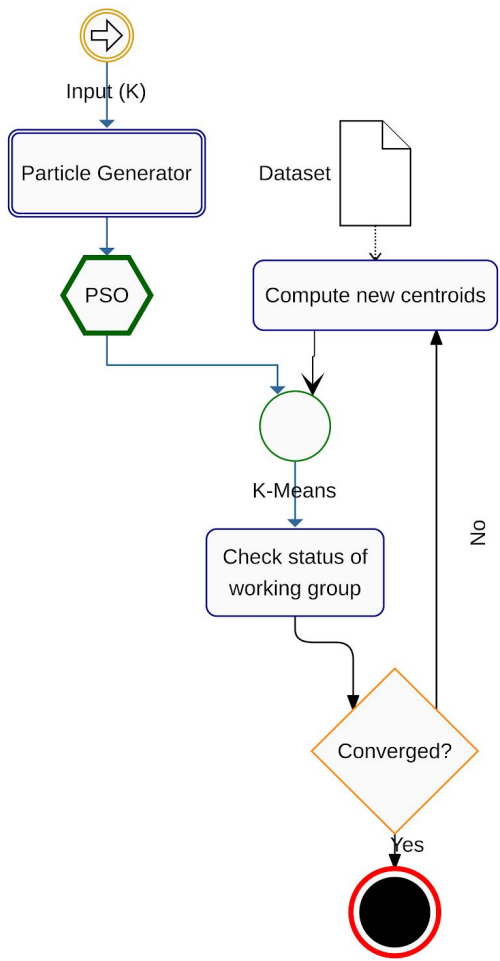

The above diagram illustrates the overall idea. Once the value for 'k' is provided, several sample particles are randomly initialized. These particles then form the swarm that would be further used for optimization.

The fitness function of these particles are the loss values given by the following formula.

$$f_a = \frac{\sqrt{\sum_S (x_c - x)^2}}{N}$$

Here, xc and x are the respective dimensional coordinates of the centroids and the data points respectively. The following section illustrates some snapshots of the implementation for the same.

### III. EXPERIMENTS

Here are some snapshots illustrating our implementation of the proposed techniques. For the purpose of this paper, I have used the famous Iris dataset, since it is very globular in nature, thus suitable for K-means.

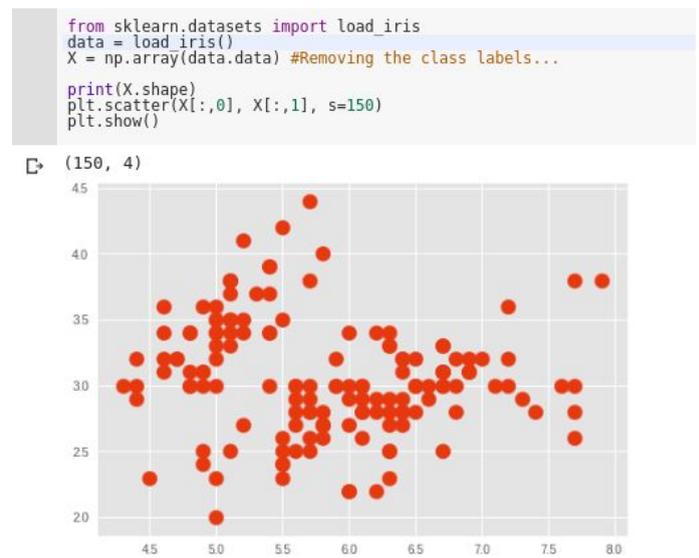

I imported the dataset from the publicly available Scikit-learn library. This would ensure uniform and standard results.

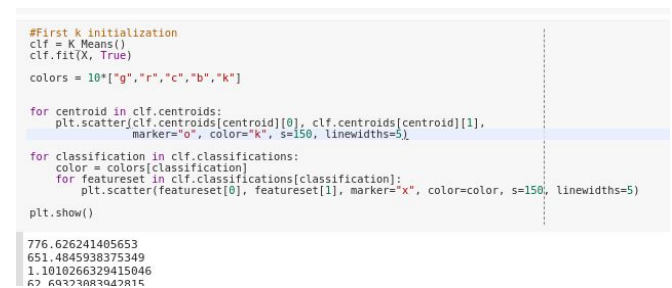

Here is a normal implementation of K-means, with randomly initialized centroids, taken straight from scikit-learn. **It took 24 iterations for it to converge.**

I understand that the code might not be visible, but the code is hosted publicly on Github. Feel free to experiment with the code. Any contributions and suggestions are welcome.

Since I could not use the Sklearn implementation to work with PSO, I implemented the vanilla K-means from scratch using Python. I also implemented PSO, partly inspired by YeswantKeswani's Genetic K-means implementation on Github. He follows a similar approach, but instead of PSO, he used Genetic Algorithm for the optimization task.

I've hosted the code on my github account, ashupednekar. Here are the results of the PSO optimized K-means. Note: I took 4 as my testing value for 'k'.

```
Particle Swarm Optimisation

PARAMETERS
---------
Population size : 100
Dimensions      : 4
Error Criterion : 1e-05
c1              : 2
c2              : 2
function        : f6
RESULTS
-------
gbest fitness : 0.9627759248925122
gbest params  : [5.39179612 3.21420312 2.27593784 0.50551819]
iterations    : 10001
[5.3917804  3.21422944 2.27587572 0.50549549]
The optimized position is, [5, 3, 2, 1]
```

```
#PSO initialization
clf = K_Means()
clf.fit_pso_ini(X, good_pos, True)

colors = 10*["g","r","c","b","k"]
```

```
994.28314060667
1710.257129499766
1189.3604234860884
1428.6224489795916
8.088235294117645
```

This took just five iterations for the clustering to converge with similar loss metrics. I've plotted both the cluster formations for reference.

IV. RESULTS

The following plot shows the clusters formed by the normal K-means implementation.

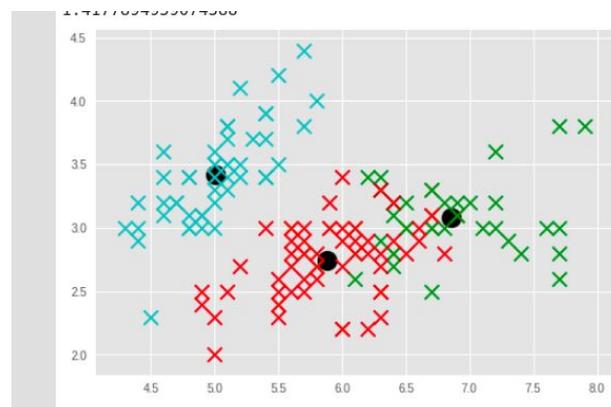

Now let's see what the PSO optimized K-means could achieve. It is important to note that this result was obtained in just five iterations, unlike the former implementation that took 24 iterations. Here's the resultant plot.

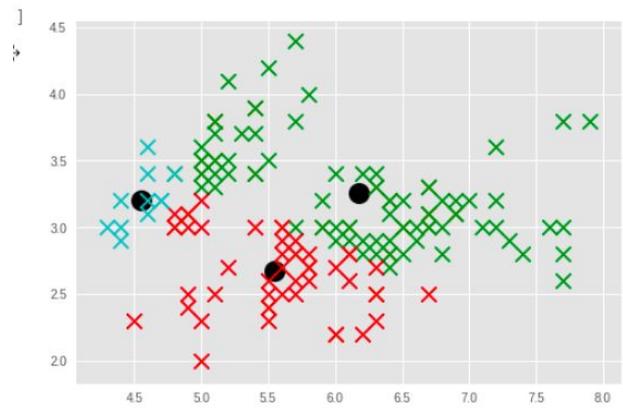

So it can be clearly seen that this approach shows great promise as it obtained acceptable results in around **six times** less number of iterations.

ACKNOWLEDGMENT *(Heading 5)*

I would like to express my solemn gratitude to my University professors as well as sources like Udacity and Coursera for guiding me towards machine learning research in general. That is what made this paper possible. I would also like to thank the awesome open source community on GitHub for their inspiration and support.

REFERENCES

[1] Krishna, K., and Narasimha M. Murty. "Genetic K-means algorithm." IEEE Transactions on Systems Man And Cybernetics-Part B: Cybernetics 29.3 (1999): 433-439.

[2] Wagstaff, Kiri, et al. "Constrained k-means clustering with background knowledge." Icml. Vol. 1. 2001.

[3] Kennedy, James. "Swarm intelligence." Handbook of nature-inspired and innovative computing. Springer, Boston, MA, 2006. 187-219.

[4] Bonabeau, E., Marco, D. D. R. D. F., Dorigo, M., Théraulaz, G., & Theraulaz, G. (1999). Swarm intelligence: from natural to artificial systems (No. 1). Oxford university press.

[5] Kennedy, James. "Particle swarm optimization." Encyclopedia of machine learning (2010): 760-766.

[6] Muthiah-Nakarajan, Venkataraman, and Mathew Mithra Noel. "Galactic Swarm Optimization: A new global optimization metaheuristic inspired by galactic motion." Applied Soft Computing 38 (2016): 771-787.

[7] Niknam, T., & Amiri, B. (2010). An efficient hybrid approach based on PSO, ACO and k-means for cluster analysis. Applied soft computing, 10(1), 183-197.

[8] Ahmadyfard, A., & Modares, H. (2008, August). Combining PSO and k-means to enhance data clustering. In 2008 International Symposium on Telecommunications (pp. 688-691). IEEE.

[9] Sculley, D., 2010, April. Web-scale k-means clustering. In Proceedings of the 19th international conference on World wide web (pp. 1177-1178). ACM.

[10] Davidson, Ian, and S. S. Ravi. "Clustering with constraints: Feasibility issues and the k-means algorithm." In Proceedings of the 2005 SIAM international conference on data mining, pp. 138-149. Society for Industrial and Applied Mathematics, 2005.